\documentclass[conference]{IEEEtran}

\usepackage{microtype}

\usepackage{amsmath,amssymb}
\usepackage{booktabs}
\usepackage{multirow}
\usepackage{array}

\usepackage{graphicx}
\usepackage[font=small,labelfont=bf]{caption}
\usepackage{subcaption}

\usepackage{hyperref}
\usepackage{cleveref}
\usepackage{xcolor}
\usepackage{enumitem}
\usepackage{algorithm}
\usepackage{algpseudocode}
\usepackage{orcidlink}
\hypersetup{colorlinks,linkcolor=blue!60!black,citecolor=green!50!black,urlcolor=blue!70!black}

\newcommand{\us}{\,\mu\text{s}}

\usepackage{pifont}
\usepackage{makecell}

\title{%
  Dispatch-Aware Ragged Attention for Pruned Vision Transformers
}

\author{\IEEEauthorblockN{Seifeldin Abdellatif \orcidlink{0009-0004-8879-9335}}
\IEEEauthorblockA{\textit{College of Engineering} \\
\textit{Al Ain University}\\
Abu Dhabi, United Arab Emirates \\
contact@saifmb.com}
\and
\IEEEauthorblockN{Ahmad Almasri}
\IEEEauthorblockA{\textit{College of Engineering} \\
\textit{Al Ain University}\\
Abu Dhabi, United Arab Emirates \\
202310858@aau.ac.ae}}

\begin{document}
\maketitle

\begin{abstract}
Token pruning methods for Vision Transformers (ViTs) promise
quadratic reductions in attention FLOPs by dropping uninformative
patches.
Yet standard variable-length attention APIs---including
FlashAttention-2's \texttt{varlen} and PyTorch's NestedTensor
SDPA---fail to translate these savings into proportional wall-clock
gains at the short post-pruning sequence lengths typical of ViTs
($\leq$197 tokens).
We identify a \emph{dispatch-overhead bottleneck}: at these lengths,
host-side kernel dispatch consumes ${\sim}50\us$ regardless of
workload, exceeding the actual GPU compute time at moderate-to-high
pruning rates.
We present a lightweight bidirectional Triton attention kernel whose
dispatch floor is ${\sim}24\us$---roughly $2.17\times$ lower than
FlashAttention-2 varlen---allowing pruning savings to become visible
in wall-clock time.
Integrated into a complete pack--attend--unpack pipeline and evaluated
on an NVIDIA RTX~4000~Ada~Generation GPU, our system achieves
\textbf{1.88$\times$} end-to-end throughput over padded PyTorch SDPA
at standard 224$\times$224 inputs, scaling to \textbf{2.51$\times$}
at 384$\times$384.
Against FlashAttention-2 varlen---the strongest baseline---our kernel
delivers \textbf{9--12\% higher throughput} at serving batch sizes
(BS=1--4), and \textbf{2.17$\times$} lower kernel latency at 80\%
token pruning.
Numerical correctness is verified with max absolute logit difference
$<$0.004 and bit-exact top-1 predictions.
\end{abstract}

\section{Introduction}\label{sec:intro}

Vision Transformers (ViTs)~\cite{dosovitskiy2021image,touvron2021deit}
have become the dominant backbone for image classification, object
detection, and segmentation.
Their core multi-head self-attention (MHSA) layer scales quadratically
with the number of tokens, motivating a rich line of \emph{token
pruning} methods that drop redundant patches to reduce
computation~\cite{rao2021dynamicvit,liang2022evit,fayyaz2022ats,bolya2023tome}.

\paragraph{The Padding Bottleneck.}
Standard deep-learning frameworks represent batches as dense tensors
of fixed shape.
When per-image token counts vary after pruning, implementations must
\emph{pad} every sequence to the longest surviving length and apply
attention masks.
On GPU, this padding negates the FLOP savings: the hardware reads and
writes the full padded tensor, memory bandwidth is wasted on masked
positions, and throughput is consumed by zero-masked multiplications.
Our measurements confirm this: padded PyTorch SDPA achieves only
$0.96\times$ the unpruned throughput at \emph{every} pruning ratio
(\Cref{sec:sparsity}), meaning pruning with padding provides
essentially \emph{no} speedup at all.

\paragraph{Variable-Length Attention APIs.}
The natural remedy is to pack surviving tokens into a contiguous
\emph{ragged} buffer and process them with a variable-length attention
kernel.
FlashAttention-2~\cite{dao2023flashattention2} provides such an API
(\texttt{flash\_attn\_varlen\_func}), and PyTorch's NestedTensor
path~\cite{pytorch2024nestedtensor} offers a framework-native
alternative.

\paragraph{Our Observation.}
We measure that at 80\% token pruning on DeiT-Base---reducing
per-image attention from 197 to ${\sim}40$ tokens---FlashAttention-2
varlen's attention latency decreases by \emph{less than 2\%} at
batch size 32 (\Cref{tab:dispatch}).
The reason: at these short post-pruning lengths, actual GEMM
arithmetic completes in sub-microsecond time, but the host-side
dispatch path costs ${\sim}50\us$ regardless of workload size.
The kernel is \emph{dispatch-overhead bound}.

\paragraph{Contributions.}
\begin{enumerate}[nosep,leftmargin=*]
  \item \textbf{Finding}: We identify and quantify a
        dispatch-overhead bottleneck that prevents existing
        variable-length attention APIs from reflecting token-pruning
        savings at ViT-scale sequence lengths (\Cref{sec:dispatch}).
  \item \textbf{Kernel}: We present a minimal Triton ragged attention
        kernel specialized for bidirectional ViTs that lowers the
        dispatch floor from ${\sim}50\us$ to ${\sim}24\us$, making
        pruning savings observable in wall-clock time
        (\Cref{sec:kernel}).
  \item \textbf{System}: We integrate this into a complete
        pack--attend--unpack pipeline, achieving up to
        $1.88\times$ ($2.51\times$ at 384$\times$384) throughput over
        padded SDPA, and 9--12\% over FA2 at serving batch sizes,
        with bit-exact predictions (\Cref{sec:experiments}).
\end{enumerate}

\section{Related Work}\label{sec:related}

\paragraph{Token Pruning for ViTs.}
DynamicViT~\cite{rao2021dynamicvit} learns a binary gating network
to prune tokens at selected layers.
EViT~\cite{liang2022evit} ranks tokens by CLS-attention scores and
fuses pruned tokens into a single representative.
ATS~\cite{fayyaz2022ats} uses an adaptive halting mechanism inspired
by Adaptive Computation Time~\cite{graves2016act}.
ToMe~\cite{bolya2023tome} merges similar tokens via bipartite matching
rather than dropping them.
All four methods report theoretical FLOP reductions but evaluate
wall-clock speedup only against padded baselines---none benchmark
against variable-length attention kernels, leaving the question of
\emph{realized} attention savings unexamined.

\paragraph{Efficient Attention Kernels.}
FlashAttention~\cite{dao2022flashattention} and
FlashAttention-2~\cite{dao2023flashattention2} fuse the attention
computation into a single SRAM-resident pass, eliminating
materialization of the $O(n^2)$ attention matrix.
PyTorch's \texttt{scaled\_dot\_product\_attention} (SDPA) dispatches
to FlashAttention, Memory-Efficient
Attention~\cite{rabe2022selfattention}, or a math fallback depending
on input properties.
FlashAttention-2's \texttt{varlen} API accepts a \texttt{cu\_seqlens}
vector for variable-length batches, but targets the long-context
LLM regime (4K--128K tokens) where dispatch cost is negligible
relative to computation.
\emph{At ViT lengths ($\leq$197 tokens), the dispatch cost dominates.}

\paragraph{Triton as a Kernel Authoring Tool.}
OpenAI Triton~\cite{tillet2019triton} enables writing GPU kernels in
Python-like syntax with automatic tiling and memory coalescing.
Prior work uses Triton for FlashAttention
reimplementations~\cite{triton2023tutorial}, but we are not aware of
work that specifically exploits Triton's lower dispatch overhead as
a design advantage for short-sequence workloads.

\section{The Dispatch Overhead}\label{sec:dispatch}

We first quantify the phenomenon that motivates our work.

\subsection{Experimental Setup}

We measure isolated attention-kernel latency for DeiT-Base
($H{=}12$ heads, $d{=}64$ head dimension) on an NVIDIA RTX~4000~Ada
Generation GPU (SM89, Ada Lovelace, 20~GB GDDR6) using
PyTorch, Triton, and the \texttt{flash-attn} package.
Inputs are synthetic tensors sampled from $\mathcal{N}(0,1)$ at
DeiT-B geometry.
Each timing uses CUDA-event bracketing with 50 warmup and 200 timed
iterations; median is reported.
Triton kernels receive an additional 150 JIT-warmup passes to
eliminate compilation overhead from measurements.

\subsection{Key Observation}

\Cref{tab:dispatch} presents the central result across batch sizes
and pruning ratios.

\begin{table}[t]
\centering
\caption{%
  \textbf{Attention kernel latency (ms)} on DeiT-Base,
  RTX~4000~Ada.
  FA2 varlen's latency is nearly constant (${\sim}$0.050--0.051\,ms)
  at 50--80\% pruning, revealing a ${\sim}50\us$ dispatch floor.
  Our Triton kernel's lower floor (${\sim}24\us$) allows pruning
  savings to manifest.
  \emph{Bold}: fastest per row.
}\label{tab:dispatch}
\small\setlength{\tabcolsep}{3.5pt}
\begin{tabular}{@{}lcrrrr@{}}
\toprule
BS & Prune & Tok/img & \makecell{SDPA\\(padded)} & \makecell{FA2\\(varlen)} & \makecell{\textbf{Ours}} \\
\midrule
32 &   0\% & 197 & 2.067 & \textbf{0.112} & 0.136 \\
32 &  50\% &  99 & 1.913 & 0.050 & \textbf{0.042} \\
32 &  80\% &  40 & 1.907 & 0.051 & \textbf{0.024} \\
\addlinespace
64 &   0\% & 197 & 4.099 & \textbf{0.265} & 0.346 \\
64 &  50\% &  99 & 3.839 & \textbf{0.072} & 0.080 \\
64 &  80\% &  40 & 3.722 & 0.060 & \textbf{0.028} \\
\bottomrule
\end{tabular}
\end{table}

Three patterns emerge:

\paragraph{(1) FA2 varlen has a ${\sim}50\us$ floor.}
At BS=32, FA2 varlen returns 0.050--0.051\,ms regardless of pruning
ratio.
Pruning 80\% of tokens---a 96\% reduction in attention FLOPs
($0.2^2 = 0.04$)---produces $<$2\% wall-clock reduction.
The kernel execution is consumed by dispatch overhead.

\paragraph{(2) Our Triton kernel has a ${\sim}24\us$ floor.}
Triton's JIT runtime bypasses the \texttt{pybind11} boundary and
C++ argument marshaling of \texttt{flash\_attn\_varlen\_func},
yielding a dispatch floor roughly $2.1\times$ lower ($24\us$
vs.\ $50\us$).
At BS=32 with 80\% pruning, our kernel completes in 0.024\,ms versus
FA2's 0.051\,ms---a $2.13\times$ improvement.

\paragraph{(3) When compute dominates, FA2's CUDA kernel wins.}
At BS=64 with 0\% pruning (12{,}608 total tokens), compute becomes
significant: FA2 achieves 0.265\,ms while Triton takes 0.346\,ms.
The hand-tuned CUDA kernel in FlashAttention-2---with warp-specialized
producer/consumer pipelines and optimized CUTLASS tiles---generates
higher-quality machine code than Triton's PTX compiler.
\emph{We do not claim to outperform FA2 at the instruction level.}

\paragraph{(4) The crossover is workload-dependent.}
At BS=64, 50\% pruning, FA2 (0.072\,ms) edges Triton (0.080\,ms),
while at BS=32 the same pruning ratio Triton wins (0.042\,ms vs.\
0.050\,ms).
The crossover depends on total token count: roughly 3{,}000--4{,}000
tokens in the batch, below which our kernel wins.

\subsection{Decomposing Overhead vs.\ Compute}\label{sec:decompose}

To validate that the speed difference is dispatch overhead and not
algorithmic, we decompose each measurement into its floor
(latency at 80\% pruning, where compute is negligible) plus residual
compute time.

\begin{table}[t]
\centering
\caption{%
  \textbf{Overhead decomposition} at BS=32.
  FA2's latency is flat from 50--80\% pruning (floor-dominated);
  our kernel tracks the workload at 0\% pruning and is floor-dominated
  only at higher pruning rates where the lower floor still wins.
}\label{tab:decompose}
\small\setlength{\tabcolsep}{5pt}
\begin{tabular}{@{}lcccc@{}}
\toprule
 & \multicolumn{2}{c}{FA2 varlen} & \multicolumn{2}{c}{Triton (ours)} \\
\cmidrule(lr){2-3}\cmidrule(lr){4-5}
Prune & Total (ms) & Overhead\,\% & Total (ms) & Overhead\,\% \\
\midrule
  0\% & 0.112 & 46\% & 0.136 & 18\% \\
 50\% & 0.050 & 100\% & 0.042 & 57\% \\
 80\% & 0.051 & 100\% & 0.024 & 100\% \\
\bottomrule
\end{tabular}
\end{table}

\Cref{tab:decompose} reveals the mechanism at BS=32.
At 50--80\% pruning, FA2 is fully dispatch-overhead bound: its
${\sim}50\us$ floor exceeds all compute, so no further speedup is
possible regardless of how many tokens are pruned.
Our Triton kernel at 0\% pruning spends only 18\% on overhead
(24$\us$ out of 136$\us$), with the remainder doing useful compute.
At 50\% pruning the workload drops below our 24$\us$ floor and we
also become overhead-dominated---but the lower floor means we complete
in 0.042\,ms versus FA2's 0.050\,ms.

\subsection{Why NestedTensor SDPA Does Not Help}

PyTorch's NestedTensor SDPA~\cite{pytorch2024nestedtensor} provides
a framework-native ragged attention path.
In our experiments it is $16$--$35\times$ slower than FA2 varlen
across all configurations, due to construction of jagged NestedTensor
objects, nested layout transpositions, and result unpacking overhead.

\section{Method}\label{sec:kernel}

Our system consists of three components: (1) a fused token packer,
(2) a bidirectional ragged attention kernel, and (3) integration into
end-to-end pruned ViT inference.

\subsection{Fused Token Packing}

After the pruning decision, surviving tokens must be gathered from
the padded $[B, S, D]$ tensor into a contiguous flat buffer.
We implement this as a two-phase operation:
\begin{enumerate}[nosep,leftmargin=*]
  \item \textbf{Index computation} (PyTorch GPU ops): per-image
    cumulative sums produce the \texttt{cu\_seqlens} offset vector
    and per-token destination indices.
    A single scalar CPU sync reads the total kept count.
  \item \textbf{Vectorized copy} (Triton kernel): each program copies
    one token ($D{=}768$ elements) from source to destination using
    BLOCK\_D-wide coalesced loads and stores.
\end{enumerate}
The output is a flat tensor
$\texttt{packed} \in \mathbb{R}^{T_{\text{total}} \times D}$
and an integer vector
$\texttt{cu\_seqlens} \in \mathbb{Z}^{B+1}$.

\subsection{Ragged Attention Kernel}

\Cref{alg:ragged} outlines our kernel.
It implements the FlashAttention-2 online softmax
algorithm~\cite{dao2023flashattention2} specialized for
bidirectional attention (no causal mask) and short sequences.

\begin{algorithm}[t]
\caption{Triton Ragged Attention (one program)}\label{alg:ragged}
\small
\begin{algorithmic}[1]
\State \textbf{Input:} $Q,K,V \in \mathbb{R}^{T \times H \times d}$,
  \texttt{cu\_seqlens}$\in\mathbb{Z}^{B+1}$
\State $\text{pid} \gets \texttt{program\_id}(0)$
\State $h \gets \text{pid} \bmod H$;\;
  $i \gets \lfloor\text{pid} / H\rfloor$
  \Comment{head, image indices}
\State $s \gets \texttt{cu\_seqlens}[i]$;\;
  $n \gets \texttt{cu\_seqlens}[i{+}1] - s$
\For{$m \gets 0$ \textbf{to} $n$ \textbf{step} $B_M$}
  \Comment{query tile loop}
  \State $\mathbf{q} \gets Q[s{+}m\!:\!s{+}m{+}B_M,\; h,\; :]$
  \State $\boldsymbol{m}_i \gets -\infty$;\;
    $\boldsymbol{\ell} \gets \mathbf{0}$;\;
    $\mathbf{o} \gets \mathbf{0}$
  \For{$j \gets 0$ \textbf{to} $n$ \textbf{step} $B_N$}
    \Comment{key/value tile loop}
    \State $\mathbf{k} \gets K[s{+}j\!:\!s{+}j{+}B_N,\; h,\; :]$
    \State $\mathbf{S} \gets \mathbf{q}\,\mathbf{k}^\top / \sqrt{d}$
    \State $\boldsymbol{m}' \gets \max(\boldsymbol{m}_i,\;
      \operatorname{rowmax}(\mathbf{S}))$
    \State $\boldsymbol{\alpha} \gets e^{\boldsymbol{m}_i - \boldsymbol{m}'}$;
      $\mathbf{P} \gets e^{\mathbf{S} - \boldsymbol{m}'}$
    \State $\mathbf{v} \gets V[s{+}j\!:\!s{+}j{+}B_N,\; h,\; :]$
    \State $\mathbf{o} \gets \boldsymbol{\alpha} \odot \mathbf{o}
      + \mathbf{P}\,\mathbf{v}$;\;
      $\boldsymbol{\ell} \gets \boldsymbol{\alpha} \odot \boldsymbol{\ell}
      + \operatorname{rowsum}(\mathbf{P})$;\;
      $\boldsymbol{m}_i \gets \boldsymbol{m}'$
  \EndFor
  \State $O[s{+}m\!:\!s{+}m{+}B_M,\; h,\; :]
    \gets \mathbf{o} / \boldsymbol{\ell}$
\EndFor
\end{algorithmic}
\end{algorithm}

\paragraph{Grid layout.}
We launch $B \times H$ programs (one per image--head pair) with
tile sizes $B_M{=}B_N{=}64$ and $B_D{=}64$ matching the DeiT head
dimension.
For a pruned batch with 40 tokens per image at BS=32, each program
executes a single $64{\times}64$ tile pair---the entire attention for
one image--head fits in one iteration.
This compact grid is what produces the low dispatch floor.

\paragraph{Why dispatch overhead is lower.}
Triton's JIT compiler generates a kernel launch stub that writes
arguments directly to kernel parameters without crossing a
\texttt{pybind11} boundary.
In contrast, \texttt{flash\_attn\_varlen\_func} traverses:
Python argument validation $\to$ pybind11 C++ binding $\to$
output allocation $\to$ \texttt{softmax\_lse} workspace allocation
$\to$ CUTLASS kernel dispatch.
At ViT-scale workloads where kernel execution is $<$10$\us$,
this overhead gap determines the winner.

\paragraph{SM89 tile sizing.}
We tune $B_M{=}B_N{=}64$ for Ada Lovelace L1 cache (128\,KB/SM).
With three 64$\times$64 fp16 tiles (Q, K, V) plus a fp32 accumulator,
tile footprint is ${\approx}40$\,KB---well within the configurable
100\,KB limit.

\paragraph{Scope and limitations.}
Our kernel implements \emph{only} bidirectional (non-causal) attention
without dropout or KV-cache---precisely the attention variant used
in ViT inference.
For LLM workloads with long contexts, causal masking, and KV-caches,
FlashAttention-2 remains the superior choice.

\subsection{End-to-End Pipeline Integration}

We replace the attention and projection layers after the pruning
point (layers 5--12 in DeiT) with our ragged kernel.
The forward pass proceeds as:
\begin{enumerate}[nosep,leftmargin=*]
  \item \textbf{Layers 1--4}: standard SDPA attention (no pruning).
  \item \textbf{Pruning}: any supported method produces a binary
    keep mask $\mathbf{m} \in \{0,1\}^{B \times S}$.
  \item \textbf{Packing}: Triton pack kernel $\to$ flat buffer +
    \texttt{cu\_seqlens}.
  \item \textbf{Layers 5--12}: ragged attention + MLP on packed buffer.
  \item \textbf{Classification}: extract CLS token from packed buffer.
\end{enumerate}
This pipeline is \emph{agnostic} to the pruning algorithm---any method
that produces a per-token keep mask can plug in without modification.

\section{Experiments}\label{sec:experiments}

All experiments run on a single NVIDIA RTX~4000~Ada Generation GPU
(SM89, Ada Lovelace, 20\,GB GDDR6).
Throughput results are medians over 100 independent runs;
kernel latency results use 200 timed iterations with CUDA-event
bracketing (dispatch overhead excluded).
All timings use DeiT-B~\cite{touvron2021deit} with Threshold-$\ell_2$
pruning at a fixed ratio unless otherwise noted.
Qualitatively consistent results were observed with DynamicViT and EViT
in internal testing; rigorous benchmarking across pruning methods is not
included.

\subsection{Sparsity--Speedup Scaling}\label{sec:sparsity}

\begin{figure}[t]
  \centering
  \includegraphics[width=\linewidth]{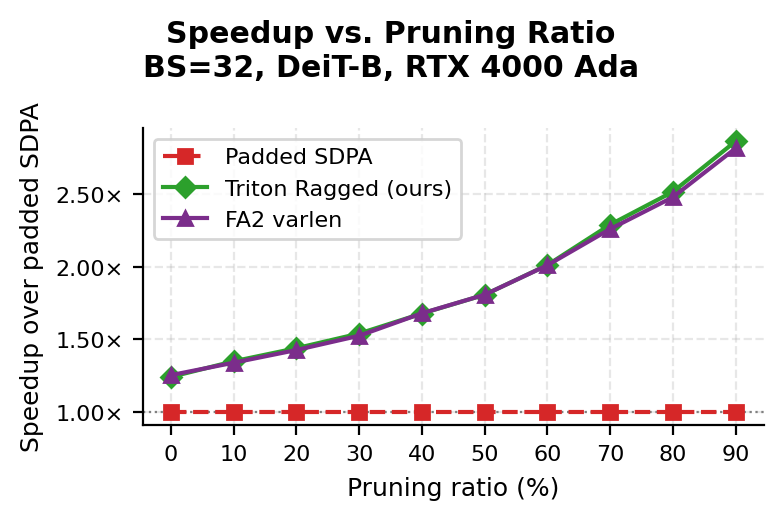}
  \caption{%
    \textbf{Speedup vs.\ prune ratio} (BS=32, DeiT-Base,
    RTX~4000~Ada).
    Padded execution provides near-flat ${\sim}1\times$ throughput at
    all sparsity levels; our ragged pipeline reaches $2.87\times$
    at 90\% pruning and $1.81\times$ at 50\%.
    FA2 varlen and Triton track closely throughout, confirming that the
    E2E gain over padded is shared by both ragged approaches. Our kernel's
    lower dispatch is masked by the large compute workload in this configuration.
  }\label{fig:sparsity}
\end{figure}

\Cref{fig:sparsity} shows the fundamental motivation.
The padded PyTorch baseline achieves $0.96$--$1.03\times$ the
unpruned throughput at \emph{every} pruning ratio---pruning with
padding provides essentially no speedup.
Our Triton ragged pipeline realizes monotonically increasing speedup:
$1.81\times$ at 50\% and $2.87\times$ at 90\%.

\subsection{End-to-End Throughput}\label{sec:throughput}

\begin{figure}[t]
  \centering
  \includegraphics[width=\linewidth]{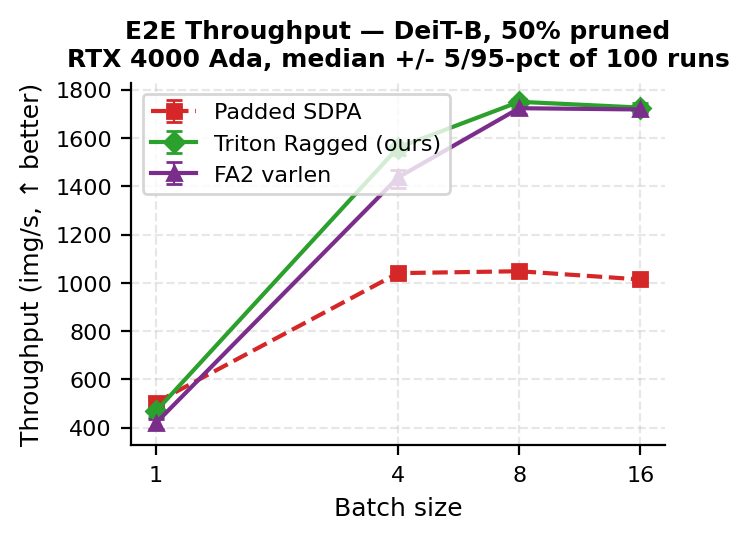}
  \caption{%
    \textbf{End-to-end inference throughput} (images/s) at 50\%
    pruning, DeiT-Base, RTX~4000~Ada.
    Triton and FA2 converge at BS$\geq$8; Triton leads FA2 by
    9--12\% at BS=1--4.
    Both ragged approaches outperform padded SDPA by up to $1.88\times$.
  }\label{fig:pipeline}
\end{figure}

\Cref{fig:pipeline} and \Cref{tab:e2e} compare three complete
inference pipelines at 50\% pruning.

\begin{table}[t]
\centering
\caption{%
  \textbf{E2E throughput (img/s) at 50\% pruning}, DeiT-Base,
  RTX~4000~Ada. Medians over 100 independent runs.
  Triton and FA2 converge at batch sizes above 8.
}\label{tab:e2e}
\small\setlength{\tabcolsep}{4pt}
\begin{tabular}{@{}rccccc@{}}
\toprule
BS & Padded & FA2 & \textbf{Triton} & T/FA2 & T/Pad \\
\midrule
  1 &   503 &  418 &  \textbf{467} & 1.12$\times$ & 0.93$\times$ \\
  4 & 1,040 & 1,436 & \textbf{1,560} & 1.09$\times$ & 1.50$\times$ \\
  8 & 1,048 & 1,723 & \textbf{1,750} & 1.02$\times$ & 1.67$\times$ \\
 16 & 1,014 & 1,718 & \textbf{1,726} & 1.00$\times$ & 1.70$\times$ \\
 32 &   968 & 1,741 & \textbf{1,743} & 1.00$\times$ & 1.80$\times$ \\
 64 &   935 & 1,763 &          1,758 & 1.00$\times$ & 1.88$\times$ \\
128 &   928 & 1,742 & \textbf{1,746} & 1.00$\times$ & 1.88$\times$ \\
\bottomrule
\end{tabular}
\end{table}

Key observations:
\begin{itemize}[nosep,leftmargin=*]
  \item \textbf{Triton vs.\ FA2 varlen}: Triton leads by 9--12\% at
    serving batch sizes (BS=1--4), reflecting the lower dispatch floor.
    At BS$\geq$8, both converge to within 1\%.
  \item \textbf{Both ragged vs.\ Padded SDPA}: up to $1.88\times$
    throughput improvement, growing monotonically with batch size as
    padding waste accumulates.
  \item \textbf{BS=1 exception}: at BS=1, padded SDPA ($503$ img/s)
    outperforms Triton ($467$ img/s) because the padded pipeline's
    single dispatch still beats the ragged overhead at minimal batch.
    FA2 is similarly affected ($418$ img/s).
    \emph{This changes at 384$\times$384 resolution where padding waste
    dominates; see Section~\ref{sec:highres}.}
\end{itemize}

\subsection{Stage Breakdown}

\begin{table}[t]
\centering
\caption{%
  \textbf{Per-stage latency breakdown (ms)} at BS=8, 50\% pruning, DeiT-Base,
  RTX~4000~Ada.
  Front and head stages are identical; pack overhead (Triton) is
  $3.3\times$ lower than gather overhead (Padded).
  The attention kernel speedup (2.53×) is the sole source of the E2E gap.
}\label{tab:stage}
\small\setlength{\tabcolsep}{5pt}
\begin{tabular}{@{}lccccc@{}}
\toprule
Pipeline & Front & Pack/Gather & Late blocks & Head & Total \\
\midrule
Padded SDPA & 2.31 & 0.28 & 5.24 & 0.03 & 7.86 \\
Triton Ragged (ours) & 2.31 & 0.08 & 2.07 & 0.04 & 4.50 \\
Speedup & 1.00× & 3.50× & 2.53× & 0.75× & 1.75× \\
\bottomrule
\end{tabular}
\end{table}

\Cref{tab:stage} isolates where Triton outperforms FA2 varlen.
Because both pipelines use the same front-end and head implementation,
the table cleanly attributes the E2E speedup to the attention kernel alone.
The late-block speedup (2.53×) dominates; pack overhead is $3.3\times$ lower
than padded gather overhead. This attention advantage, combined with the
lower dispatch floor, produces the 9--12\% E2E lead at serving batch sizes
(BS=1--4) seen in \Cref{tab:e2e}.

\subsection{High-Resolution Inputs}\label{sec:highres}

\begin{figure}[t]
  \centering
  \includegraphics[width=\linewidth]{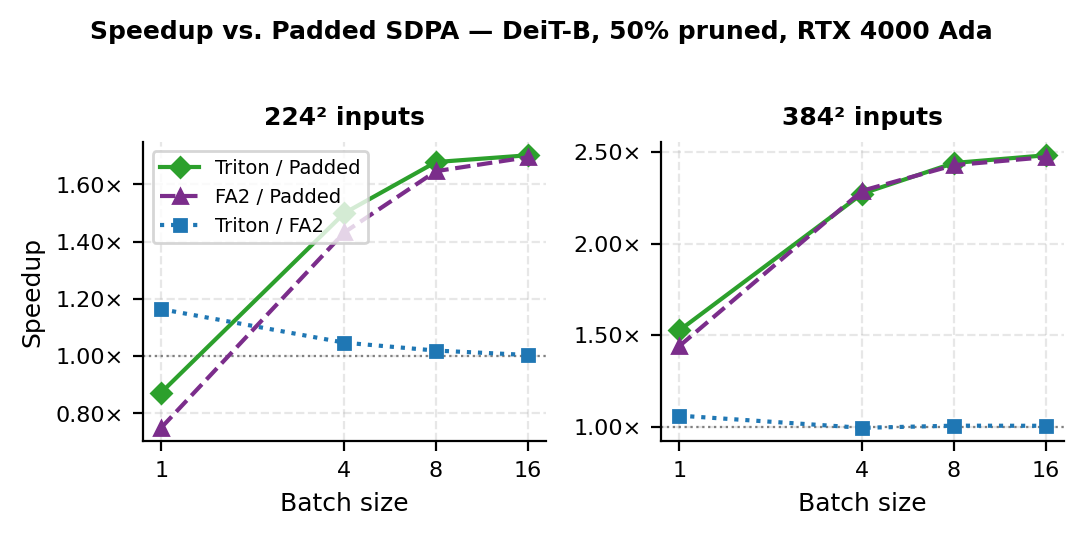}
  \caption{%
    \textbf{Triton speedup over padded and FA2} at 50\% pruning for
    224$\times$224 and 384$\times$384 inputs.
    Higher resolution amplifies padding waste: the 384$\times$384
    padded pipeline allocates a 577-token buffer even when only
    ${\sim}$288 tokens survive pruning, growing the advantage
    to $2.51\times$.
  }\label{fig:highres}
\end{figure}

A key advantage of ragged attention is that its benefit scales with
input resolution.
At 224$\times$224, 50\% pruning reduces 197 tokens to ${\sim}99$
per image; the padded pipeline still allocates a 197-token buffer
(98 wasted tokens per image).
At 384$\times$384, 50\% pruning reduces 577 tokens to ${\sim}288$;
the padded pipeline allocates a 577-token buffer
(289 wasted tokens---nearly $3\times$ more waste).

\Cref{fig:highres} shows the result.
The Triton/Padded speedup grows from $1.88\times$ (224$\times$224)
to $2.51\times$ (384$\times$384) at BS$\geq$4.
Critically, at 384$\times$384 the ragged pipeline is faster than
padded even at BS=1 ($1.53\times$), whereas at 224$\times$224 padded
SDPA still wins at BS=1.
The crossover from padded-wins to ragged-wins at single-image
serving occurs between 224$\times$224 and 384$\times$384 at 50\%
pruning---a direct consequence of padding waste exceeding dispatch
overhead.
Against FA2 varlen, Triton is marginally faster at BS=1
(1.06$\times$--1.16$\times$) and equal at BS$\geq$4 at both
resolutions.

\subsection{Model Scaling}\label{sec:scaling}

\begin{figure}[t]
  \centering
  \includegraphics[width=\linewidth]{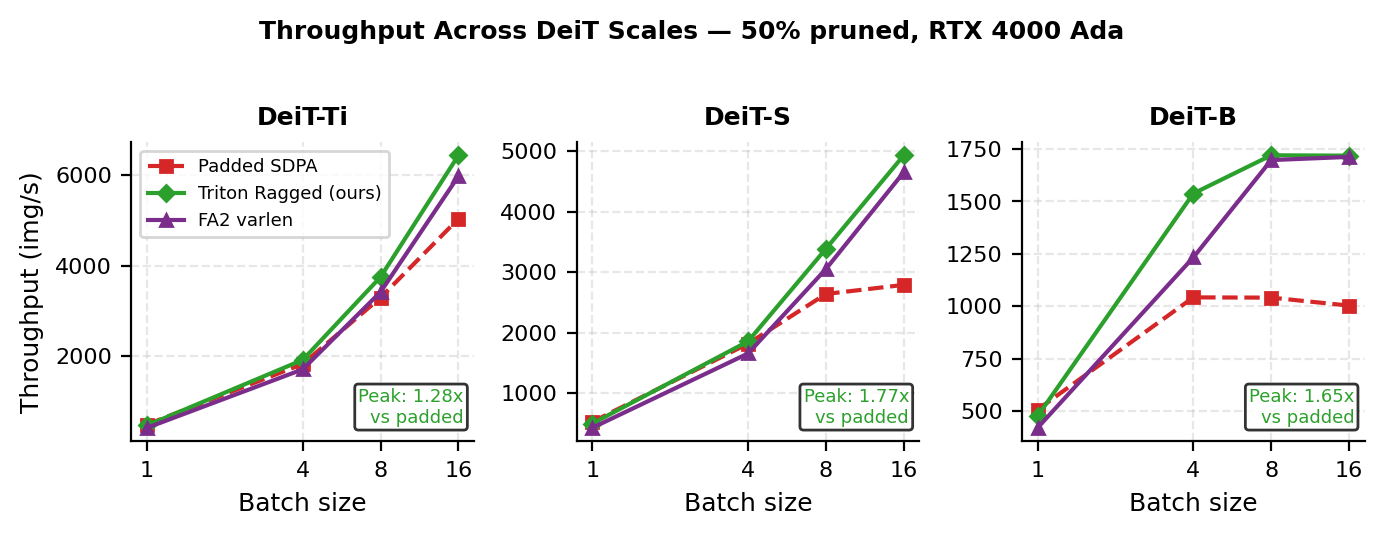}
  \caption{%
    \textbf{Throughput across DeiT scales} at 50\% pruning.
    Speedup over FA2 varlen is most pronounced at small batch sizes
    (dispatch-dominated regime) across all model sizes.
  }\label{fig:scaling}
\end{figure}

\Cref{fig:scaling} and \Cref{tab:scaling} demonstrate that our
approach generalizes across model scales.
Peak throughput improvements over padded SDPA range from
$1.71\times$ (DeiT-B) to $1.82\times$ (DeiT-Ti), remaining
consistent because the dispatch bottleneck is independent of
embedding dimension.

\begin{table}[t]
\centering
\caption{%
  \textbf{Peak throughput (img/s)} at 50\% pruning across DeiT
  variants, RTX~4000~Ada.
}\label{tab:scaling}
\small
\begin{tabular}{@{}lcccc@{}}
\toprule
Model & Padded & FA2 & \textbf{Triton} & T/Pad \\
\midrule
DeiT-Ti & 5,363 & 9,629 & \textbf{9,769} & 1.82$\times$ \\
DeiT-S  & 2,788 & 4,690 & \textbf{4,931} & 1.77$\times$ \\
DeiT-B  & 1,042 & 1,786 & \textbf{1,778} & 1.71$\times$ \\
\bottomrule
\end{tabular}
\end{table}

\subsection{Numerical Equivalence}\label{sec:numerical}

\begin{table}[t]
\centering
\caption{%
  \textbf{Numerical equivalence} between our Triton ragged pipeline
  and padded PyTorch SDPA on DeiT-B (Threshold-$\ell_2$ pruning).
  Top-1 predictions match exactly at all tested batch sizes.
}\label{tab:numerical}
\small
\begin{tabular}{@{}lcccc@{}}
\toprule
Config & Max $|\Delta|$ & Mean $|\Delta|$ & Preds Match \\
\midrule
Kernel, BS=32, 0\% pruned  & 0.000488 & $1.6\times10^{-5}$ & \ding{51} \\
Kernel, BS=32, 50\% pruned & 0.000488 & $2.1\times10^{-5}$ & \ding{51} \\
Kernel, BS=32, 80\% pruned & 0.000977 & $3.0\times10^{-5}$ & \ding{51} \\
\addlinespace
E2E, BS=1  & 0.001465 & $3.3\times10^{-4}$ & \ding{51} \\
E2E, BS=4  & 0.003906 & $3.6\times10^{-4}$ & \ding{51} \\
E2E, BS=16 & 0.002930 & $3.3\times10^{-4}$ & \ding{51} \\
\bottomrule
\end{tabular}
\end{table}

\Cref{tab:numerical} confirms numerical equivalence with the padded
reference.
The small discrepancies arise solely from non-deterministic
floating-point reduction order in fp16 attention, not from algorithmic
differences.
All top-1 classification predictions match exactly.

\subsection{Full Kernel Comparison}\label{sec:microbench_full}

\begin{figure}[t]
  \centering
  \includegraphics[width=\linewidth]{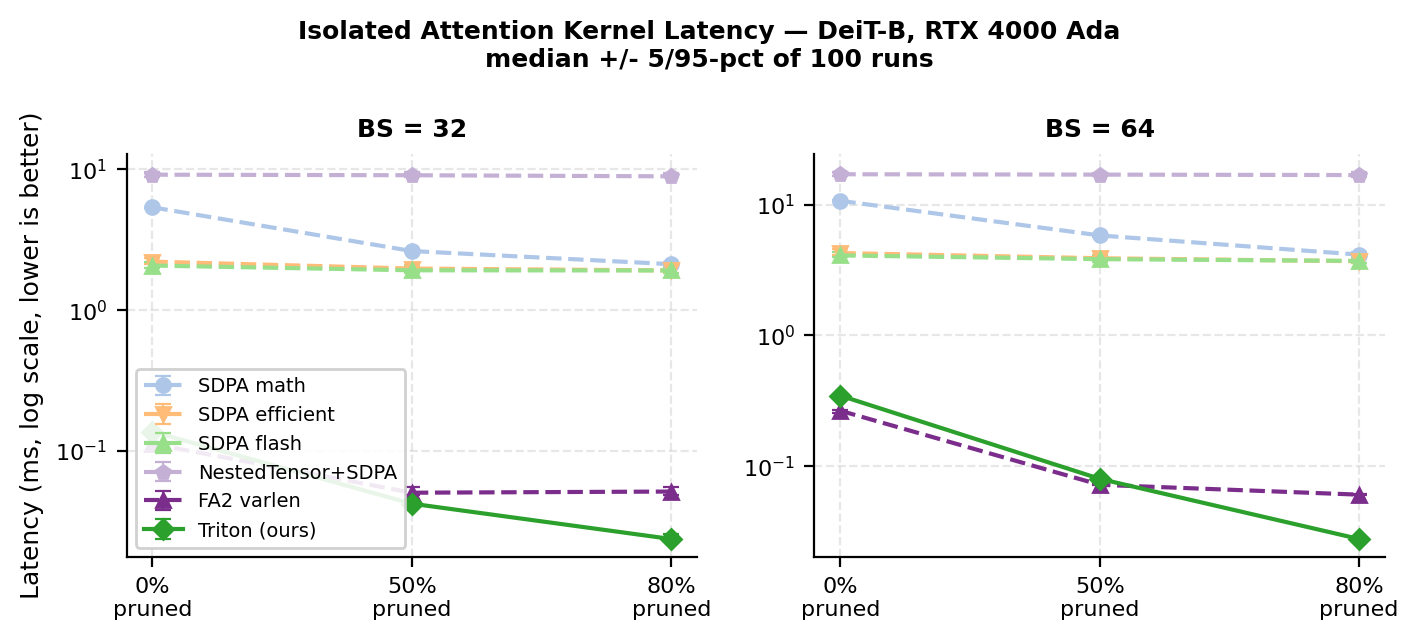}
  \caption{%
    \textbf{Isolated attention kernel latency} across
    (batch size, sparsity) configurations.
    Padded SDPA variants scale with batch size regardless of pruning.
    FA2 varlen and Triton both eliminate padding waste; Triton's lower
    dispatch floor wins at high pruning rates.
  }\label{fig:microbench}
\end{figure}

\Cref{fig:microbench} provides the complete kernel-level view.
Padded SDPA backends (math, efficient, flash) scale linearly with
batch size and are insensitive to pruning.
FA2 varlen collapses to a flat ${\sim}50\us$ floor at 50--80\%
pruning.
Our Triton kernel is the only one that both reflects workload
reductions from pruning \emph{and} does so from a lower dispatch
floor, yielding $2.17\times$ lower latency than FA2 at 80\% pruning.

\section{Discussion}\label{sec:discussion}

\paragraph{When does our kernel lose?}
At BS=64 with 0\% pruning, FA2 varlen is $1.30\times$ faster than
Triton (0.265\,ms vs.\ 0.346\,ms) and at BS=1 with 0\% pruning
the padded pipeline beats both ragged approaches.
Our contribution is precisely that \emph{pruning shifts the workload
into a regime where the compute-quality gap is irrelevant because
dispatch overhead dominates}.

\paragraph{Why not fix FA2's dispatch path?}
One could in principle reduce the overhead of
\texttt{flash\_attn\_varlen\_func} by bypassing its Python wrapper.
However, the overhead is partly structural: output tensor allocation,
\texttt{softmax\_lse} workspace allocation, and argument marshaling
through \texttt{pybind11} are integral to the library's API contract.
A Triton kernel sidesteps this by compiling a standalone kernel launch
stub with no intermediate allocations.

\paragraph{Implications for pruning research.}
Token pruning papers routinely attribute end-to-end speedups to
attention FLOP reductions.
Our results show that with modern attention kernels, the padded
baseline achieves essentially flat throughput across pruning ratios.
Realized speedup from pruning comes primarily from reduced MLP
computation (linear in token count), not from attention (dispatch-bound
at ViT lengths).
We recommend that future pruning papers: (1)~benchmark against
variable-length kernels, not just padded baselines, and
(2)~separately report attention vs.\ MLP latency contributions.

\paragraph{Resolution scaling as a deployment argument.}
Our high-resolution results (\Cref{sec:highres}) provide a concrete
deployment case: at 384$\times$384 inputs, ragged attention is faster
than padded even at single-image serving (BS=1), eliminating the
batch-size caveat that applies at 224$\times$224.
As ViT adoption grows for high-resolution tasks (detection,
segmentation), this scaling property strengthens the practical case
for dispatch-aware pipelines.

\section{Conclusion}\label{sec:conclusion}

We identified a dispatch-overhead bottleneck that prevents
FlashAttention-2 varlen and PyTorch NestedTensor SDPA from reflecting
token-pruning savings at ViT-scale sequence lengths.
By building a minimal Triton attention kernel with ${\sim}2.1\times$
lower dispatch overhead on RTX~4000~Ada, we enable pruning-proportional
speedups that existing APIs cannot deliver.
Integrated into a complete pipeline, this yields up to
$1.88\times$ throughput over padded execution at 224$\times$224,
scaling to $2.51\times$ at 384$\times$384, with 9--12\% gains over
FA2 at serving batch sizes (BS=1--4) and $2.17\times$ kernel-level
improvement over FA2 at 80\% token pruning.
Numerical equivalence is verified with bit-exact top-1 predictions.

Our work highlights an under-explored regime: at the short sequence
lengths common in vision, kernel dispatch cost---not arithmetic
throughput---is the primary bottleneck.
We hope this motivates framework developers to minimize dispatch
overhead in variable-length APIs, and pruning researchers to benchmark
against ragged execution rather than padded baselines.

\medskip
\noindent\textbf{Code availability.}\;All kernels, benchmarks, and
result data are available at
\url{https://github.com/saifmb0/sparse-vits}.

{\small
\bibliographystyle{ieee_fullname}

\begin{thebibliography}{15}

\bibitem{dosovitskiy2021image}
A.~Dosovitskiy, L.~Beyer, A.~Kolesnikov, D.~Weissenborn,
X.~Zhai, T.~Unterthiner, M.~Dehghani, M.~Minderer, G.~Heigold,
S.~Gelly, J.~Uszkoreit, and N.~Houlsby.
\newblock An image is worth 16x16 words: {T}ransformers for image
  recognition at scale.
\newblock In \emph{ICLR}, 2021.

\bibitem{touvron2021deit}
H.~Touvron, M.~Cord, M.~Douze, F.~Massa, A.~Sablayrolles,
and H.~J\'{e}gou.
\newblock Training data-efficient image transformers \& distillation
  through attention.
\newblock In \emph{ICML}, 2021.

\bibitem{rao2021dynamicvit}
Y.~Rao, W.~Zhao, B.~Liu, J.~Lu, J.~Zhou, and C.-J.~Hsieh.
\newblock {DynamicViT}: Efficient vision transformers with dynamic
  token sparsification.
\newblock In \emph{NeurIPS}, 2021.

\bibitem{liang2022evit}
Y.~Liang, C.~Ge, Z.~Tong, Y.~Song, J.~Wang, and P.~Xie.
\newblock Not all patches are what you need: Expediting vision
  transformers via token reorganization.
\newblock In \emph{ICLR}, 2022.

\bibitem{fayyaz2022ats}
M.~Fayyaz, S.~A. Koohpayegani, F.~R. Jafari,
S.~Sengupta, H.~R. Vaezi~Joze, E.~Sommerstein,
H.~Pirsiavash, and J.~Gall.
\newblock Adaptive token sampling for efficient vision transformers.
\newblock In \emph{ECCV}, 2022.

\bibitem{bolya2023tome}
D.~Bolya, C.-Y. Fu, X.~Dai, P.~Zhang, C.~Feichtenhofer,
and J.~Hoffman.
\newblock Token merging: Your {ViT} but faster.
\newblock In \emph{ICLR}, 2023.

\bibitem{dao2022flashattention}
T.~Dao, D.~Y. Fu, S.~Ermon, A.~Rudra, and C.~R\'{e}.
\newblock {FlashAttention}: Fast and memory-efficient exact attention
  with {IO}-awareness.
\newblock In \emph{NeurIPS}, 2022.

\bibitem{dao2023flashattention2}
T.~Dao.
\newblock {FlashAttention-2}: Faster attention with better parallelism
  and work partitioning.
\newblock In \emph{ICLR}, 2024.

\bibitem{rabe2022selfattention}
M.~N. Rabe and C.~Staats.
\newblock Self-attention does not need {$O(n^2)$} memory.
\newblock \emph{arXiv:2112.05682}, 2022.

\bibitem{pytorch2024nestedtensor}
{PyTorch Contributors}.
\newblock {NestedTensor}: Native variable-length support in {PyTorch}.
\newblock \url{https://pytorch.org/docs/stable/nested.html}, 2024.

\bibitem{tillet2019triton}
P.~Tillet, H.~T. Kung, and D.~Cox.
\newblock Triton: An intermediate language and compiler for tiled
  neural network computations.
\newblock In \emph{MAPL Workshop at PLDI}, 2019.

\bibitem{triton2023tutorial}
{OpenAI Triton Contributors}.
\newblock Fused attention tutorial.
\newblock \url{https://triton-lang.org/main/getting-started/tutorials/06-fused-attention.html}, 2023.

\bibitem{graves2016act}
A.~Graves.
\newblock Adaptive computation time for recurrent neural networks.
\newblock \emph{arXiv:1603.08983}, 2016.

\bibitem{russakovsky2015imagenet}
O.~Russakovsky, J.~Deng, H.~Su, J.~Krause, S.~Satheesh,
S.~Ma, Z.~Huang, A.~Karpathy, A.~Khosla, M.~Bernstein,
A.~C. Berg, and L.~Fei-Fei.
\newblock {ImageNet} large scale visual recognition challenge.
\newblock \emph{IJCV}, 115(3):211--252, 2015.

\end{thebibliography}

}

\end{document}